\documentclass{myifcolog}
\usepackage{xcolor}
\usepackage{url}
\usepackage{amsmath}
\usepackage{amsthm}
\usepackage{amsfonts}
\usepackage{xspace}
\usepackage{graphicx}
\usepackage{caption,subcaption}
\usepackage[textsize=tiny]{todonotes}
\usepackage{pdflscape}
\usepackage{afterpage}
\usepackage{capt-of}
\usepackage[final]{pdfpages}
\newcommand{\lif}[0]{
  \leftarrow
}
\newcommand{\rif}[0]{
  \rightarrow
}
\newcommand{\vt}[1]{
\mathbf{#1}
}

\newcommand{\fzand}[0]{
  \wedge
}

\newcommand{\pr}[1]{
  \mathrm{#1}
}

\begin{document}


\title{Neural-Symbolic Computing: An Effective  Methodology for Principled Integration of Machine Learning and Reasoning}
\titlerunning{Neural-symbolic computing}

\addauthor[a.garcez@city.ac.uk, mxgori@gmail.com, luislamb@acm.org, serafini@fbk.eu, michael.spranger@gmail.com, sn.tran@utas.edu.au]{Artur d'Avila Garcez, Marco Gori, Luis C. Lamb, Luciano Serafini, Michael Spranger, Son N. Tran\thanks{Corresponding author. Authors are in alphabetical order.}}{City, Univ. of London, Univ. of Siena, UFRGS, FBK, Sony Japan, Univ. of Tasmania}

\authorrunning{Garcez et al.}

\titlethanks{We thank Richard Evans for his valuable comments and suggestions.}

\maketitle

\begin{abstract}
Current advances in Artificial Intelligence and machine learning in general, and deep learning in particular have reached unprecedented impact not only across research communities, but also over popular media channels. However, concerns about interpretability and accountability of AI have been raised by influential thinkers. In spite of the recent impact of AI, several works have identified the need for principled knowledge representation and reasoning mechanisms integrated with deep learning-based systems to provide sound and explainable models for such systems. Neural-symbolic computing aims at integrating, as foreseen by Valiant, two most fundamental cognitive abilities: the ability to learn from the environment, and the ability to reason from what has been learned. Neural-symbolic computing has been an active topic of research for many years, reconciling the advantages of robust learning in neural networks and reasoning and interpretability of symbolic representation. In this paper, we survey  recent accomplishments of neural-symbolic computing as a principled methodology for integrated machine learning and reasoning. We illustrate the effectiveness of the approach  by outlining the main characteristics of the methodology: principled integration of neural learning with symbolic knowledge representation and reasoning allowing for the construction of  explainable AI systems. The insights provided by neural-symbolic computing shed new light on the increasingly prominent need for interpretable and accountable AI systems.
\end{abstract}
\section{Introduction}
Current advances in Artificial Intelligence (AI) and machine learning in general, and deep learning in particular have reached unprecedented impact not only within the academic and industrial research communities, but also among popular media channels. Deep learning researchers have achieved groundbreaking results and built AI systems that have in effect rendered new paradigms in areas such as computer vision, game
playing, and natural language processing 
\cite{Hinton-nature,Silver_2017}. 
Nonetheless, the impact of deep learning has been so remarkable that 
leading entrepreneurs such as Elon Musk and Bill Gates, and outstanding scientists such as Stephen Hawking have voiced strong concerns about AI's accountability, impact on humanity and even on the future of the planet \cite{nature-ethics}.

Against this backdrop, researchers have recognised the need for offering a better understanding of the underlying principles of AI systems, in particular those based on machine learning, aiming at establishing solid foundations for the field. In this respect, Turing Award Winner Leslie Valiant had already pointed out that one of the key challenges for AI in the coming decades is the development of integrated reasoning and learning mechanisms, so as to construct a rich semantics of intelligent cognitive behavior \cite{Valiant2003}. In Valiant's words:
\emph{``The aim here is to identify a way of looking at and manipulating commonsense
knowledge  that  is  consistent  with  and  can  support  what  we  consider  to  be  the
two most fundamental aspects of intelligent cognitive behavior: the ability to learn
from experience, and the ability to reason from what has been learned. We are
therefore seeking
a semantics of knowledge that can computationally support the
basic phenomena of intelligent behavior."} 
In order to respond to these scientific, technological and societal challenges which demand reliable, accountable and explainable AI systems and tools, the integration of cognitive abilities ought to be carried out in a principled way.

Neural-symbolic computing aims at integrating, as put forward by Valiant, two most fundamental cognitive abilities: the ability to learn from  experience, and the ability to reason from what has been learned \cite{Bader_2005,Ivan_2017,Garcez_2008}. 
The integration of learning and reasoning through neural-symbolic computing has been an active branch of AI research for several years \cite{Evans_18,Garcez_2008,Garcez_2002,hammer-hitzler,Khardon_97,Serafini_2016,Valiant_2006}. Neural-symbolic computing aims at reconciling the dominating symbolic and connectionist paradigms of AI under a principled foundation. 
In neural-symbolic computing, knowledge is represented in symbolic form, whereas learning and reasoning are computed by a neural network. Thus, the underlying characteristics of neural-symbolic computing allow the principled combination of robust learning and efficient inference in neural networks, along with interpretability offered by symbolic knowledge extraction and reasoning with logical systems. 
  
Importantly, as AI systems started to outperform humans in certain tasks \cite{Silver_2017}, several ethical and societal concerns were raised \cite{nature-ethics}. Therefore, the interpretability and explainability of AI systems become crucial alongside their accountability.

In this paper, we survey the principles of neural-symbolic integration by highlighting key characteristics that underline this research paradigm. 
Despite their differences, both the symbolic and connectionist paradigms, share common characteristics offering benefits when integrated in a principled way (see e.g. \cite{NIPS03,Garcez_2008,Smolensky_1995,Valiant_2006}).
For instance, neural learning and inference under uncertainty may address the brittleness of symbolic systems. On the other hand, symbolism provides additional knowledge for learning  which may e.g. ameliorate neural network's well-known catastrophic forgetting or difficulty with extrapolating. 
In addition, the integration of neural models with logic-based symbolic models provides an AI system capable of bridging lower-level information processing (for perception and pattern recognition) and higher-level abstract knowledge (for reasoning and explanation).

In what follows, we review the important and recent developments of research on neural-symbolic systems. 
We start by outlining the main important characteristics of a neural-symbolic system: Representation, Extraction, Reasoning and Learning \cite{Bader_2005,Garcez_2002}, and their applications. We then discuss and categorise the approaches to representing symbolic knowledge in neural-symbolic systems into three main groups: rule-based, formula-based and embedding-based. After that, we show the capabilities and applications of neural-symbolic systems for learning, reasoning, and explainability. Towards the end of the paper we outline recent trends and identify a few challenges for neural-symbolic computing research.



\section{Prolegomenon to Neural-Symbolic Computing} 
Neural-symbolic systems have been applied successfully to several fields, including data science, ontology learning, training and assessment in simulators, and models of cognitive learning and reasoning \cite{Borges_2011,Evans_18,Garcez_2008,Leo_2011}. 
However, the recent impact of deep learning in vision and language processing and the growing complexity of 
(autonomous) AI systems demand improved explainability and accountability. 
In neural-symbolic computing, learning, reasoning and knowledge extraction are combined. Neural-symbolic systems are modular and seek to have the property of compositionality. This is achieved through the streamlined representation of several knowledge representation languages which are computed by connectionist models. The
{\it Knowledge-Based Artificial Neural Network} (KBANN) \cite{Towel_1994} and the {\it Connectionist inductive learning and logic programming} (CILP) \cite{Garcez_2002} systems were some of the most influential models that combine logical reasoning and neural learning. 
As pointed out in \cite{Garcez_2002} KBANN served as inspiration in the construction of the CILP
system. CILP provides a sound theoretical foundation to inductive learning and reasoning in artificial neural networks through theorems showing how logic programming can be a knowledge representation language for neural networks. 
The KBANN system was the first to allow for learning with background knowledge in neural networks and knowledge extraction, with relevant applications in bioinformatics. CILP allowed for the integration of learning, reasoning and knowledge extraction in recurrent networks. 
An important result of CILP was to show how neural networks endowed with semi-linear neurons approximate the fixed-point operator of propositional logic programs with negation. This result allowed applications of reasoning and learning using backpropagation and logic programs as background knowledge \cite{Garcez_2002}. 

Notwithstanding, the need for richer cognitive models soon demanded the representation and learning of other forms of reasoning, such as temporal reasoning, reasoning about uncertainty, epistemic, constructive and argumentative reasoning \cite{Garcez_2008,Valiant2003}. Modal and temporal logic have achieved first class status in the formal toolboxes of AI and Computer Science researchers. In AI, modal logics are amongst the most widely used logics in the analysis and modelling of reasoning in distributed multiagent systems.
In the early 2000s, researchers then showed that ensembles of CILP neural networks, when properly set up, can compute the \emph{modal} fixed-point operator of modal and temporal logic programs. In addition to these results, such ensembles of neural networks were shown to represent the possible world semantics of modal propositional logic, fragments of first order logic and of linear temporal logics. 
In order to illustrate the computational power of \emph{Connectionist Modal Logics (CML)} and \emph{Connectionist Temporal Logics of Knowledge (CTLK)} \cite{NIPS03,NECO3050}, researchers were able to learn full solutions to several problems in distributed, multiagent learning and reasoning, including the Muddy Children Puzzle \cite{NIPS03} and the Dining Philosophers Problem \cite{AAAI07}. 

\begin{figure}[h]
\centering
\includegraphics[width=0.6\textwidth]{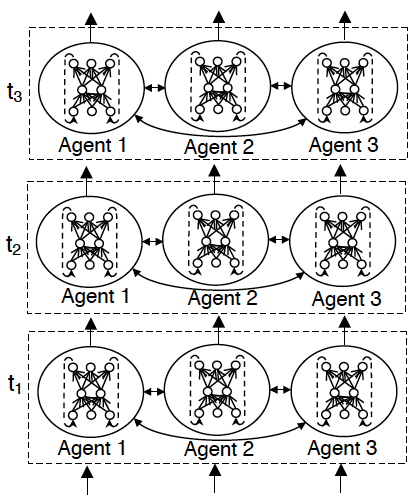}
\caption{Evolution of Reasoning and Learning in Time}
\label{fig:temporal}
\end{figure}

By combining temporal logic with modalities, one can represent knowledge and learning evolution in time. This is a key insight, allowing for temporal evolution of both learning and reasoning in time (see Fig. \ref{fig:temporal}). The Figure represents the integrated learning and reasoning process of CTLK. At each time point (or one state of affairs), e.g. $t_{2}$, knowledge which the agents are endowed with and what the agents have learned at the previous time $t_{1}$ is represented. As time progresses, linear evolution of the agents' knowledge is represented in time as more knowledge about the world (what has been learned) is represented. Fig. \ref{fig:temporal} illustrates this dynamic property of CTLK, which allows not only the analysis of the current state of affairs but also of how knowledge and learning evolve over time.

Modal and temporal reasoning, when integrated with connectionist learning provide neural-symbolic systems with richer knowledge representation languages and better interpretability. As can be seen in Fig. \ref{fig:temporal}, they enable the construction of more modular deep networks. As argued by Valiant, the construction of cognitive models integrating rich logic-based knowledge representation languages, with robust learning algorithms provide an effective alternative to the construction of semantically sound cognitive neural computational models. It is also argued that a language for describing the algorithms of deep neural networks is needed. Non-classical logics such as logic programming in the context of neuro-symbolic systems, and functional languages used in the context of probabilistic programming are two prominent candidates. In the coming sections, we explain how neural-symbolic systems can be constructed from simple definitions which underline the streamlined integration of knowledge representation, learning, and reasoning in a unified model.

\section{Knowledge Representation in Neural Networks}
Knowledge representation is the cornerstone of a neural-symbolic
system that provides a mapping mechanism between symbolism and
connectionism, where logical calculus can be carried out exactly or
approximately by a neural network. This way, given a trained neural network,
symbolic knowledge can be extracted for explaining and reasoning purposes. The representation approaches can be categorised into three
main groups: rule-based, formula-based and embedding, which are
discussed as follows.
\subsection{Propositional Logic}
\label{rep:proposition}
\subsubsection{Rule-based Representation}
\label{subsubsec:rule_based}
\begin{figure}[ht]
  \centering
  \begin{subfigure}{0.45\textwidth}
    \centering
    \includegraphics[width=0.5\textwidth]{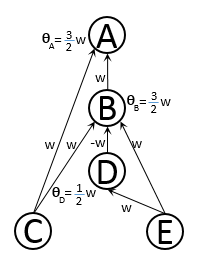}
    \caption{KBANN ($\theta$ denotes a threshold).}
    \label{kbann}
  \end{subfigure}
  \begin{subfigure}{0.45\textwidth}
    \centering
    \includegraphics[width=0.6\textwidth]{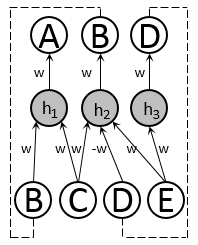}
    \caption{CILP.}
    \label{cilp}
  \end{subfigure}
  \caption{Knowledge representation of $\phi = \{\pr{A} \lif \pr{B} \fzand \pr{C}, \pr{B} \lif \pr{C} \fzand \neg \pr{D}\fzand \pr{E}, \pr{D} \lif \pr{E}\}$ using KBANN and CILP.}
\end{figure}
Early work on representation of symbolic knowledge in connectionist
networks focused on tailoring the models' parameters to establish an
equivalence between input-output mapping function of artificial neural
networks (ANN) and logical inference rules. It has been shown that by
constraining the weights of a neural network, inference
with feedforward propagation can exactly imitate the behaviour of
modus ponens \cite{Towel_1994,Garcez_1999}. KBANN \cite{Towel_1994}
employs stack of perceptrons to represent the inference rule of
logical implications. For example, given a set of rules:
\begin{equation}
  \label{kb1}
  \phi = \{\pr{A} \lif \pr{B} \fzand \pr{C}, \pr{B} \lif \pr{C} \fzand \neg \pr{D}\fzand \pr{E}, \pr{D} \lif \pr{E}\}
\end{equation}

an ANN can be constructed as in Figure \ref{kbann}. CILP then generalises the
idea by using recurrent networks and bounded continuous units
\cite{Garcez_1999}. This representation method allows the use of various data types and more complex sets of rules. With CILP, knowledge given in
Eq. \eqref{kb1} can be encoded in a neural network as shown in Figure
\ref{cilp}. In order to adapt this system to first-order logic, CILP++
\cite{Franca_2014} makes use of techniques from Inductive Logic Programming (ILP). In CILP++, examples and background knowledge are converted into propositional clauses by a {bottom-clause propositionalisation} technique, which are then encoded into an ANN with recurrent connections as done by CILP. 

\subsubsection{Formula-based Representation}
\begin{figure}[ht]
  \centering
  \begin{subfigure}{0.45\textwidth}
    \centering
    \includegraphics[width=.95\textwidth]{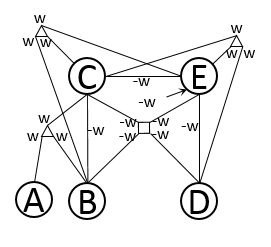}
    \caption{Higher-order network for Penalty Logic.}
    \label{penalty}
  \end{subfigure}
  \begin{subfigure}{0.5\textwidth}
    \centering
    \includegraphics[width=.95\textwidth]{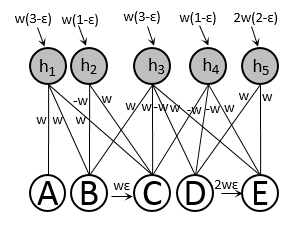}
    \caption{RBM with confidence rules.}
    \label{confidence}
  \end{subfigure}
  \caption{Knowledge representation of $  \phi = \{w:\pr{A} \lif \pr{B} \fzand \pr{C}, w:\pr{B} \lif \pr{C} \fzand \neg \pr{D}\fzand \pr{E}, w:\pr{D} \lif \pr{E}\}$ using Penalty logic and Confidence rules}
\end{figure}
One issue with KBANN-style rule-based representations is that the discriminative structure of ANNs will only allow a subset of the variables (the
consequent of the if-then formula) to be inferred, unless recurrent networks are deployed, with the other variables (the antecedents) being seen as inputs only. This would not represent the
behaviour of logical formulas and does not support general
reasoning where any variable can be inferred. In order to solve this
issue, generative neural networks can be employed as they can treat
all variables as non-discriminative. In this formula-based approach, typically associated with restricted Boltzmann machines (RBMs) as a building block, the focus is on mapping logical formulas to symmetric connectionist networks, each
characterised by an energy function. Early work such as penalty logic
\cite{Pinkas_1995} proposes a mechanism to represent weighted formulas
in energy-based connectionist (Hopfield) networks where maximising satisfiability
is equivalent to minimising energy function. Suppose that each
formula in the knowledge base \eqref{kb1} is assigned a weight $w$.
Penalty logic constructs a higher-order Hopfield network as shown 
in Figure \ref{penalty}. However, inference with such type of network is difficult, while converting the higher-order energy
function to a quadratic form is possible but computationally
expensive. Recent work on confidence rules \cite{Son_2018} proposes an
efficient method to represent propositional formulas in restricted
Boltzmann machines and deep belief networks where inference and learning
become easier. Figure \ref{confidence} shows an RBM for the knowledge
base \eqref{kb1}. Nevertheless, learning and reasoning with
restricted Boltzmann machines are still complex, making it more difficult to apply formula-based representations than rule-based
representations in practice. The main issue has to do with the
partition functions of symmetric connectionist networks which cannot
be computed analytically. This intractability problem, fortunately,
can be ameliorated using sum-product approach as has been shown
in \cite{poon_2011}. However, it is not yet clear how to apply this idea to RBMs.

\subsection{First-order Logic}
\subsubsection{Propositionalisation}
\label{subsubsec:propositionalisation}
Representation of knowledge in first-order logic in neural networks has been an ongoing challenge, but it can benefit from studies of propositional logic representation \ref{rep:proposition} using propositionalisation techniques \cite{Muggleton_1995}. Such techniques allow a first-order knowledge base to be converted into a propositional knowledge base so as to preserve entailment. In neural-symbolic computing, bottom clause prositionalisation (BCP) is a popular approach because bottom clause literals can be encoded directly into neural networks as data features while at the same time presenting semantic meaning. 

Early work from \cite{DiMaio_2004} employs prositionalisation and feedforward neural networks to learn a clause evaluation function which helps improve the efficiency in exploring large hypothesis spaces. In this approach, the neural network does not work as a standalone ILP system, instead it is used to approximate clause evaluation scores to decide the direction of the hypothesis search. In \cite{Pitangui_2012}, prositionalisation is used for learning first-
order logic in Bayesian networks. Inspired by this work, in \cite{Franca_2014}, the CILP++ system is proposed by integrating bottom clauses and rule-based approach CILP \cite{Garcez_2002}, referred to in Section \ref{subsubsec:rule_based}. 

The main advantage of propositionalisation is that it is efficient and it fits neural networks well. Also, it does not require first-order formulas to be provided as bottom clauses. However, propositionalisation has serious disadvantages. First, with function symbols, there are infinitely many ground terms. Second, propositionalization seems to generate lots of irrelevant clauses.
\subsubsection{Tensorisation}
\begin{figure}[ht]
  \centering
  \includegraphics[width=0.7\textwidth]{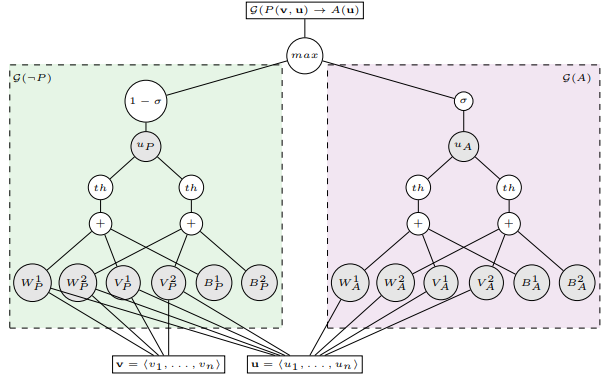}
  \caption{Logic tensor network for $P(x,y) \rif A(y)$ with
    $\mathcal{G}(x)=\vt{v}$ and $\mathcal{G}(y)=\vt{u}$; $\mathcal{G}$
    are grounding (vector representation) for symbols in first-order
    language; and the tensor order in this example is $2$ \cite{Serafini_2016}.}
  \label{ltn}
\end{figure}
Tensorisation is a class of approaches that embeds first-order logic symbols such as constants, facts and rules into real-valued tensors. Normally, constants are represented as one-hot vectors (first order tensor). Predicates and functions are matrices (second-order tensor) or higher-order tensors.

In early work, embedding techniques were
proposed to transform symbolic representations into vector spaces where
reasoning can be done through matrix computation
\cite{Bordes_2011,Socher_2013,Ilya_2008,Serafini_2016,Santoro_2017,Cohen_2017, Evans_18, Yang_2017, Dong_2019,Rocktaschel_2016}. Training
embedding systems can be carried out as distance learning using
backpropagation. Most research in this direction focuses on
representing relational predicates in a neural network. This is known
as "relational embedding"
\cite{Bordes_2011,Santoro_2017,Socher_2013,Ilya_2008}. For representation
of more complex logical structures, i.e. first order-logic formulas, a
system named {\it Logic Tensor Network} (LTN) \cite{Serafini_2016} is
proposed by extending {\it Neural Tensor Networks} (NTN)
\cite{Socher_2013}, a state-of-the-art relational embedding
method. Figure \ref{ltn} shows an example of LTN for $P(x,y) \rif
A(y)$. Related ideas are discussed formally in the context of 
constraint-based learning and reasoning \cite{Marco_book}. Recent research in first-order logic programs has successfully exploited the advantages of distributed representations of logic symbols for efficient reasoning \cite{Cohen_2017}, inductive programming  \cite{Evans_18, Yang_2017, Dong_2019}, and differentiable theorem proving \cite{Rocktaschel_2016}. 
\subsection{Temporal Logic}
One of the earliest works on temporal logic and neural networks is CTLK, where ensembles of recurrent neural networks are set up to represent the possible world semantics of linear temporal logics \cite{NIPS03}. With single hidden layers and semi-linear neurons, the networks can compute a fixed-point semantics of temporal logic rules. Another work on representation of temporal knowledge is proposed in {\it Sequential Connectionist  Temporal Logic} (SCTL) \cite{Borges_2011} where CILP is extended to work with the nonlinear auto-regressive exogenous NARX network model. {\it Neural-Symbolic Cognitive Agents} (NSCA) represent temporal knowledge in recurrent temporal RBMs \cite{Leo_2011}. Here, the temporal logic rules are modelled in the form of recursive conjunctions represented by recurrent structures of RBMs. Temporal relational knowledge embedding has been studied recently in {\it
  Tensor Product Recurrent Neural Network} (TPRN) with applications to
question-answering \cite{Hamid_2018}. 

\section{Neural-Symbolic Learning}
\subsection{Inductive Logic Programming}
Inductive logic programming (ILP) can take advantage of the learning capability of neural-symbolic computing to automatically construct a logic program from examples. Normally, approaches in ILP are categorised into {\it bottom-up} and {\it top-down} which inspire the development of neural-symbolic approaches accordingly for learning logical rules.

Bottom-up approaches construct logic programs by extracting specific clauses from examples. After that, generalisation procedures are usually applied to search for more general clauses. This is well suited to the idea of propositionalisation discussed earlier in Section \ref{subsubsec:propositionalisation}. For example, CILP++ \cite{Franca_2014} employed a bottom clause propositionalisation technique to construct CILP++. In \cite{Son_2018a}, a system called CRILP is proposed by integrating bottom clauses generated from \cite{Franca_2014} with RBMs. However, both CILP++ and CRILP learn and fine-tune formulas at a propositional level where propositionalisation would generate a large number of long clauses resulting in very large networks. This leaves an open research question of generalising bottom clauses within neural networks that scale well and can extrapolate.

Top-down approaches, on the other hand, construct logic programs from the most general clauses and extend them to be more specific. In neural-symbolic terms, the most popular idea is to take advantage of neural networks' learning and inference capabilities to fine-tune and test the quality of rules. This can be done by replacing logical operations by differentiable operations. For example, in Neural Logic Programming (NLP) \cite{Yang_2017}, learning of rules are based on the differentiable inference of TensorLog \cite{Cohen_2017}. Here, matrix computations are used to soften logic operators where the confidence of conjunctions and confidence of disjunctions are computed as product and sum, respectively. NLP generate rules from facts, starting with the most general ones. In
Differentiable Inductive Logic Programming ($\partial$ILP) \cite{Evans_18}, rules are generated from templates, which are assigned to parameters (weights) to make the loss function between actual conclusions and predicted conclusions from forward chaining differentiable.
In \cite{Rocktaschel_2016}, Neural Theorem Prover (NTP) is proposed by extending the backward chaining method to be differentiable. It shows that latent predicates from rule templates can be learned through optimisation of their distributed representations. Different from \cite{Yang_2017, Evans_18, Rocktaschel_2016} where clauses are generated and then softened by neural networks, in Neural Logic Machines (NLM) \cite{Dong_2019} the relation of predicates is learned by a neural network where input tensors represent facts (predicates of different arities) from a knowledge base and output tensors represent new facts.
\subsection{Horizontal Hybrid Learning}
Effective techniques such as deep learning usually require large
amounts of data to exhibit statistical regularities. However, in many
cases where collecting data is difficult a small dataset would make
complex models more prone to overfitting. When prior knowledge is
provided, e.g. from domain experts, a neural-symbolic system can offer
the advantage of generality by combining logical rules/formulas with
data during learning, while at the same time using the data to
fine-tune the knowledge. It has been shown that encoding knowledge
into a neural network can result in {
  performance improvements} \cite{Garcez_1999,Ivan_2017,Towel_1994, 
  Son_2018a}. Also, it is evident that using symbolic
knowledge can help {improve the efficiency} of neural network learning \cite{Garcez_1999,Franca_2014}. Such effectiveness and
efficiency are obtained by encoding logical knowledge as controlled
parameters during the training of a model. This technique, in general terms, has been known as {learning with logical
  constraints} \cite{Marco_book}.  Besides, in the case of lacking
prior knowledge one can apply the idea of neural-symbolic integration
for {knowledge transfer learning} \cite{Son_2018}. The idea is
to extract symbolic knowledge from a related domain and transfer it to
improve the learning in another domain, starting from a network that does not necessarily have to be instilled with background knowledge. Self-transfer with
{symbolic-knowledge distillation} \cite{Hu_2016} is also useful
as it can enhance several types of deep networks such as convolutional
neural networks and recurrent neural networks. Here, symbolic
knowledge is extracted from a trained network called ``teacher'' which
then would be used to encoded as regularizers to train a ``student''
network in the same domain.

\subsection{Vertical Hybrid Learning}
Studies in neuroscience
show that some areas in the brain are used for processing input
signals e.g. visual cortices for images \cite{Grill_2004,Poggio_2016},
while other areas are responsible for logical thinking and reasoning
\cite{Shokri_2012}. Deep neural networks can learn high level
abstractions from complex input data such as images, audio, and text, which should be useful at making decisions. However, despite that
optimisation process during learning being mathematically justified, it is
difficult for humans to comprehend how a decision has been made during inference time. Therefore, placing a logic
network on top of a deep neural network to learn the relations of those
abstractions, can help the
system to be able to explain itself. In \cite{Ivan_2017}, a Fast-RCNN \cite{Girshick_2015} is used for bounding-box detection of parts of objects and on top of that, a Logic Tensor Network is used to reason about relations between parts of objects and types of such objects. In such work, the perception part (Fast-RCNN) is fixed and learning is carried out in the reasoning part (LTN). In a related approach, called DeepProbLog, end-to-end learning and reasoning have been studied \cite{Robin_2018} where outputs of neural networks are used as "neural predicates" for ProbLog \cite{DeRaedt_2007}. 

\section{Neural-symbolic Reasoning}
Reasoning is an important feature of a neural-symbolic system and has recently 
attracted much attention from the research community \cite{Evans_18}. Various attempts have
been made to perform reasoning within neural networks, both
model-based and theorem proving approaches. In neural-symbolic
integration the main focus is the integration of reasoning and
learning, so that a model-based approach is preferred. Most theorem
proving systems based on neural networks, including first-order logic
reasoning systems such as SHRUTI \cite{SHRUTI}, have been unable to perform
learning as effectively as end-to-end differentiable learning
systems. On the other hand, model-based approaches have been shown
implementable in neural networks in the case of nonmonotonic,
intuitionistic and propositional modal logic, as well as abductive reasoning and other forms of human reasoning \cite{Bader_2005,Borges_2011}. As a result, the focus of
neural-symbolic computation has changed from performing
{symbolic reasoning in neural networks}, such as for example
implementing the logical unification algorithm in a neural network, to
the {combination of learning and reasoning}, in some cases with
a much more loosely-defined approach rather than full integration,
whereby a hybrid system will contain different components which may be
neural or symbolic and which communicate with each other. 
\subsection{Forward and Backward chaining}
Forward chaining and backward chaining are two popular inference techniques for logic programs and other logical systems. In the case of neural-symbolic systems forward and backward chainings are both in general implemented by feedforward inference.

Forward chaining generates new facts from the head literals of the rules using existing facts in the knowledge base. For example, in \cite{Leo_2011}, a
``Neural-symbolic Cognitive Agent'' shows that it is possible to
perform online learning and reasoning in real-world scenarios, where
temporal knowledge can be extracted to reason about driving skills \cite{Leo_2011}. This can be seen as forward chaining over time. In $\partial$ILP \cite{Evans_18}, a differentiable function is defined for each clause to carry out a single step of forward chaining. Similar to this, NLM \cite{Dong_2019} employs neural networks as a differentiable chain for forward inference. Different from $\partial$ILP, NLM represent the outputs and inputs of neural networks as grounding tensors of predicates for existing facts and new facts respectively.

Backward chaining, on the other hand, searches backward from a goal in the knowledge base to determine whether a query is derivable or not. This form a tree search starts from the query and expands further to the literals in the body of the rules whose heads match the query. TensorLog \cite{Cohen_2017} implements backward chaining using neural networks as symbols. The idea is based on stochastic logic programs \cite{Muggleton_1996}, and soft logic is applied to transform the hypothesis search into a chain of matrix operations. In NTP, a neural system is  constructed recursively for backward chaining and unification where AND and OR operators are represented as networks. In general, backward (goal-directed) reasoning is considerably harder to achieve in neural networks than forward reasoning. This is another current line of research within neuro-symbolic computation and AI.

\subsection{Approximate Satisfiability}
Inference in the case of logic programs with arbitrary formulas is more complex. In general, one may want to search over the hypothesis space to find a solution that satisfies (mostly) the formulas and facts in the knowledge base. Exact inference, that is, reasoning maximising satisfiability, is NP-hard. For this reason, some neural-symbolic systems offer a mechanism of
{approximate satisfiability}. Tensor logic networks are trained
to approximate the best satisfiability \cite{Serafini_2016} making
inference efficient with feedforward propagation. This has made LTNs applicable successfully to the Pascal data set and image understanding
\cite{Ivan_2017}. Penalty logic shows an equivalence between minimising violation and minimising energy functions of symmetric connectionist networks \cite{Pinkas_1995}. Confidence rules, another
approximation approach, shows the relation between sampling in
restricted Boltzmann machines and search for truth-assignments which
maximise satisfiability. The use of confidence rules also allows one to measure how confident a neural network is in its own answers. Based on that, neural-symbolic system 
``confidence rule inductive logic programming (CRILP)'' was constructed and  applied to inductive logic
programming \cite{Son_2018a}.
\subsection{Relationship reasoning}
Relational embedding systems have been used for reasoning about
relationships between entities. Technically, this has been done by
searching for the answer to a query that gives the highest grounding
score \cite{Bordes_2011,Bordes_2012,Socher_2013,Ilya_2008}. Deep neural networks are also employed for visual reasoning where they learn and infer relationships and features of multiple objects in images \cite{Santoro_2017,Yi_2018,Mao_2019}.
\section{Neural-symbolic Explainability}
The (re)emergence of deep networks has again raised the question of explainability. The complex structure of a deep neural network turns 
them into a powerful learning system if one can correctly engineer its components such as type
of hidden units, regularisation and optimization methods. However,
limitations of some AI applications have heightened the need for 
explainability and interpretability of deep neural
networks. More importantly, besides improving deep neural networks for
better applications one should also look for the benefits that deep networks can
offer in terms of knowledge acquisition. 
\subsection{Knowledge Extraction}
Explainability is a promising capability of neural-symbolic systems
where the behaviour of a connectionist network can be represented in a set
of human-readable expressions. In early work, the demand for solving
``black-box'' issues of neural networks has motivated a number of
{rules extraction} methods. Most of them are discussed in the surveys \cite{Andrews_1995,Jacobsson_2005,Wang_2018}. These attempts were to search for logic rules
from a trained network based on four criteria: (a) accuracy, (b)
fidelity, (c) consistency and (d) comprehensibility
\cite{Andrews_1995}. In \cite{Garcez_2002}, a sound extraction approach
based on partially ordered sets is proposed to narrow the search of
logic rules.
However, such combinatorial approaches do not scale well
to deal with the dimensionality of current networks. As a result, gradually less attention has been paid to knowledge
extraction until recently when the combination of global and local approaches started to be investigated. The idea here is either to create modular networks with rule extraction applying to specific modules or to consider rule extraction from specific layers only. 

In \cite{Son_2013, Son_2018},
it has been shown that while {extracting conjunctive clauses}
from the first layer of a deep belief network is fast and effective,
extraction in higher layers results in a loss of accuracy. A trained
deep network can be employed instead for 
{extraction of soft-logic rules} which is less formal but more flexible \cite{Hu_2016}.
{Extraction of temporal rules} have been 
studied in \cite{Leo_2011} and generated semantic relations of domain
variables over time. Besides formal logical knowledge, hierarchical
Boolean expressions can be learned from images for object detection and
recognition \cite{Zhangzhang_2013}.
\subsection{Natural Language Generation}
For explainability purposes, another approach
couples a deep network with sequence models to {extract natural
  language knowledge} \cite{Lisa_2016}.  In \cite{Bordes_2011},
instead of investigating the parameters of a trained model,
{relational knowledge extraction} is proposed where predicates
are obtained by performing inference of a trained embedding network on
text data.
\subsection{Program Synthesis}
In the field of Program Induction, {neuro-symbolic program synthesis} (NSPS) has been proposed to construct computer programs on an incremental fashion using a large amount of input-output samples \cite{Emilio_2017}. A neural network is employed to represent partial trees in a domain-specific language are  tree nodes, symbols and rules are vector representations. Explainability can be achieved through the tree-based structure of the network. Again, this shows that the integration of neural networks and symbolic representation is indeed a solution for both scalability and explainability.
\section{Conclusions}
In this paper, we highlighted the key ideas and principles of
neural-symbolic computing. In order to do so, we illustrated the main
methodological approaches which allow for the integration of effective
neural learning with sound symbolic-based, knowledge representation
and reasoning methods.  One of the principles we highlighted in the
paper is the sound mapping between symbolic rules and neural networks
provided by neural-symbolic computing methods. This mapping allows
several knowledge representation formalisms to be used as background
knowledge for potentially large-scale learning and efficient
reasoning.  This interplay between efficient neural learning and
symbolic reasoning opens relevant possibilities towards richer
intelligent systems. The comprehensibility and compositionality of
neural-symbolic systems, offered by building networks with a logical
structure, allows for integrated learning and reasoning under
different logical systems. This opens several interesting research
lines, in which learning is endowed with the sound semantics of
diverse logics. This, in turn, contributes towards the development of
explainable and accountable AI and machine learning-based systems and
tools.

\newpage
\clearpage%
    \thispagestyle{empty}
    \begin{landscape}
        \centering 
        \includepdf[landscape=true]{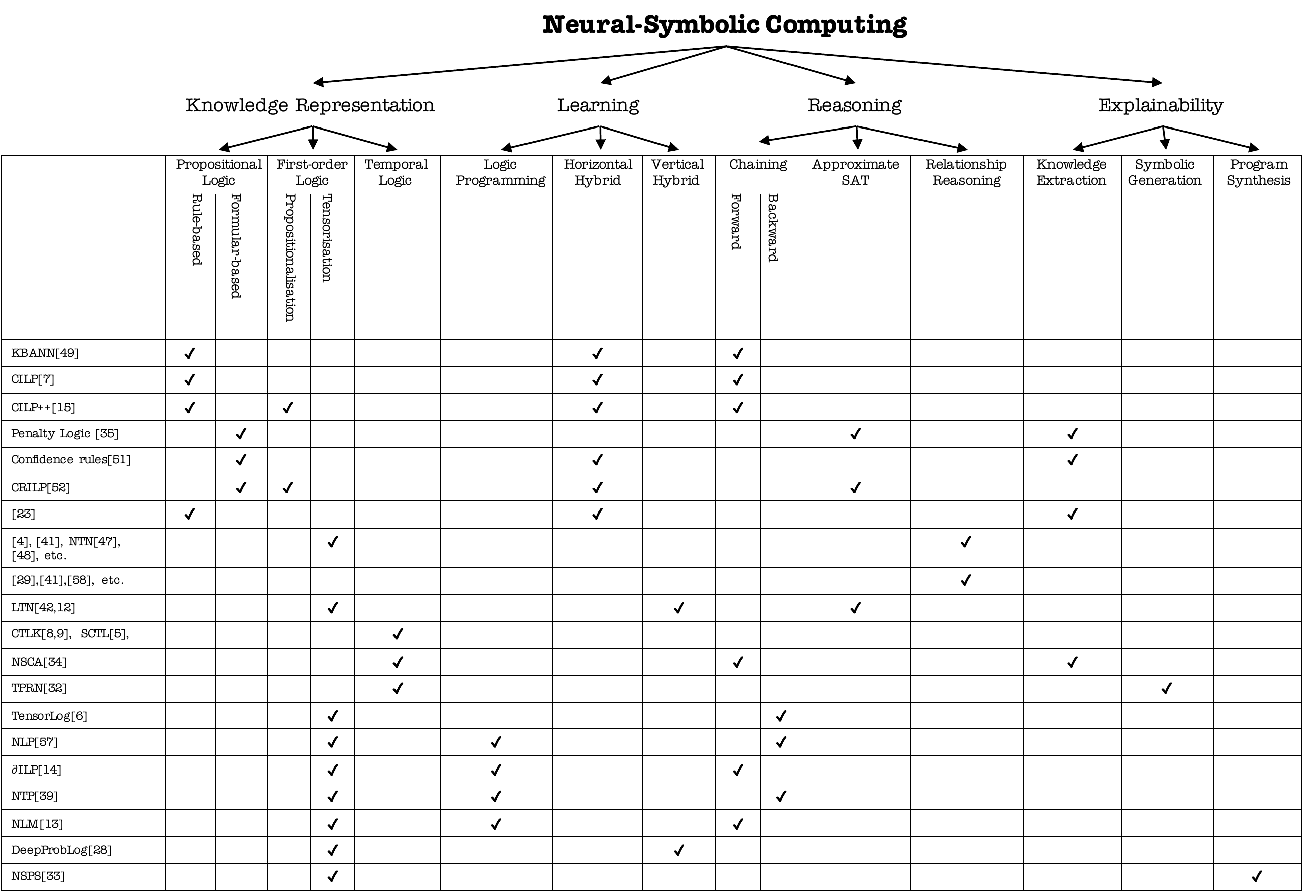}  
    \end{landscape}
    \clearpage
\small{
\bibliographystyle{plain}
\bibliography{survey}}
\end{document}